\pdfoutput=1
\documentclass[conference, 10pt]{IEEEtran}
\IEEEoverridecommandlockouts

\usepackage{color}
\usepackage{theorem}
\usepackage{times,amsmath,epsfig}
\usepackage{amssymb}
\usepackage{subcaption}
\usepackage{cite}
\usepackage{hyperref}
\usepackage{url}
\usepackage[capitalise]{cleveref}
\usepackage[ruled,linesnumbered]{algorithm2e}
\usepackage{tikz, pgfplots}
\usepackage{moresize}
\usepackage{array}

\usetikzlibrary{shapes,arrows}
\pgfplotsset{compat=1.17}
\pgfplotstableset{col sep=comma}
\usepgfplotslibrary{groupplots}

\input{my_styles.sty}

\pgfplotsset{compat=1.17}
\pgfplotstableset{col sep=semicolon}
\usepgfplotslibrary{fillbetween}
\tikzset{every mark/.append style={scale=1.5, solid}, font=\small}
\pgfplotsset{
    width=1.05\textwidth,
    height=5.5cm,
    legend style={
        font=\ssmall ,  
        inner xsep=1pt,
        inner ysep=1pt,
        nodes={inner sep=1pt}},
    legend cell align=left,
    every axis/.append style={line width=.5pt},
 	every axis plot/.append style={line width=1.5pt},
 	every axis y label/.append style={yshift=-4pt}
}
\usepgfplotslibrary{external}
\title{Structure-Guided Input Graph for GNNs facing Heterophily}
\author{{Victor M. Tenorio$^*$,
        Madeline Navarro$^\dagger$,
        Samuel Rey$^*$,
        Santiago Segarra$^\dagger$},
        and~Antonio G. Marques$^*$\\
        $^*$Dept. of Signal Theory and Communications,
        Rey Juan Carlos University, Madrid, Spain\\
        Emails: \{victor.tenorio,samuel.rey.escudero,antonio.garcia.marques\}@urjc.es \\
        $^\dagger$Dept. of Electrical and Computer Engineering, Rice University, TX, US \\
        Emails: \{nav,segarra\}@rice.edu
        
\thanks{{This work was partially supported by NSF under award CCF-2340481, Spanish AEI Grants FPU20-05554,  PID2022-136887NB-I00 and by the Autonomous Community of Madrid (course of action num. 3, Excellence Program within the multi-year agreement between CAM and URJC, and within the ELLIS Unit Madrid framework).
Research was sponsored by the Army Research Office and was accomplished under Grant Number W911NF-17-S-0002. The views and conclusions contained in this document are those of the authors and should not be interpreted as representing the official policies, either expressed or implied, of the Army Research Office or the U.S. Army or the U.S. Government. The U.S. Government is authorized to reproduce and distribute reprints for Government purposes notwithstanding any copyright notation herein.}}}

\begin{document}
%
\maketitle
\begin{abstract}
Graph Neural Networks (GNNs) have emerged as a promising tool to handle data exhibiting an irregular structure.
However, most GNN architectures perform well on homophilic datasets, where the labels of neighboring nodes are likely to be the same.
In recent years, an increasing body of work has been  devoted to the development of GNN architectures for heterophilic datasets, where labels do not exhibit this low-pass behavior.
In this work, we create a new graph in which nodes are connected if they share structural characteristics, meaning a higher chance of sharing their labels, and then use this new graph in the GNN architecture.
To do this, we compute the k-nearest neighbors graph according to distances between structural features, which are either (i) role-based, such as degree, or (ii) global, such as centrality measures.
Experiments show that the labels are smoother in this newly defined graph and that the performance of GNN architectures improves when using this alternative structure.
\end{abstract}
\begin{IEEEkeywords}
Heterophilic Datasets, Graph Neural Networks, Potential Neighbor Discovery, Graph Signal Processing, Structural Features.
\end{IEEEkeywords}

\section{Introduction}\label{s:introduction}

Graph neural networks (GNNs) have shown great success at processing data on graphs, which are assumed to encode meaningful relationships between samples~\cite{wu2021comprehensive, chien2024opportunities}.
In particular, GNNs assume nodes will be homophilic on a given graph, where nodes in closer proximity are expected to have similar features or belong to the same class~\cite{zheng2024graphneuralnetworks}.
However, while the graph structure might be informative for nodal predictions, nodal data may not follow the homophily assumption~\cite{chien2022node,suresh2021breaking,jin2020gralsp,long2020graph,lee2019graph}. 
For many real-world applications, data may exhibit heterophily, where connections often indicate greater dissimilarity between nodes~\cite{zheng2024graphneuralnetworks}. 

To address non-homophilic graph data, previous works aim to enable GNNs to incorporate relationships beyond a single-hop neighborhood.
One approach involves designing architectures that employ high-pass filtering to aggregate information from distant nodes~\cite{xu2018representation,ruiz2021graph}. 
While empirically effective, such approaches tend to be more complex and may overfit under insufficient data~\cite{rey2024redesigning}. 
An alternative approach is neighborhood discovery, where the goal is to obtain an alternative graph on which node behavior is homophilic.
This can be done by either altering the original connections or creating a new graph under additional assumptions~\cite{liu2022towards,yang2021consisrec}. 
Existing approaches such as graph transformers consider learning new connections during model training, which inevitably require large amounts of data~\cite{chien2022node}. 
Moreover, it is desirable that we understand why certain nodes are connected, but these data-driven methods may not provide explainable connections~\cite{lee2019graph,long2020graph}. 
Discovering new graph structure from data can be difficult, particularly with limited data, but under reasonable assumptions, the process might be simplified~\cite{balcilar2021analyzingexpressivepower}. 

We thus propose an intuitive neighborhood discovery approach that exploits structural characteristics.
While a given graph may be informative for predictions, it may not be suitable for GNNs that perform low-pass filtering operations~\cite{suresh2021breaking}. 
Indeed, nodal data may not be homophilic on their associated graph, but they can be related to structural characteristics~\cite{tenorio2024recovering}. 
For example, in a social network connecting professors and students at a university based on their interactions, professors are likely to have higher node degrees than students~\cite{craven1998learning}. 
Nodes are often ranked by global structural features such as centrality measures~\cite{salavati2019ranking,sheng2020identifying,segarra2015stability}. 
We thus compute two k-nearest neighbors (KNN) graphs based on each node's structural features that are either (i) role-based, describing connectivity patterns near each node, or (ii) global, determining a node's relationship to the entire graph.
Then, we consider a flexible GNN model that adaptively learns which graphs are most informative for the task at hand.
Inspired by existing findings~\cite{tenorio2024recovering,lee2019graph,long2020graph}, this approach is simpler than data-driven neighborhood discovery methods~\cite{liu2022towards,yang2021consisrec}, yet it is flexible as it adaptively combines different types of connections~\cite{peng2021reinforced}. 
Specifically, our main contributions are as follows.

\begin{table*}[t!]
    \centering
    \begin{tabular}{l|rrrr|rrrr}
    {} & \multicolumn{4}{c}{$TV (\bby)$} & \multicolumn{4}{c}{Edge homophily $h_{edge}$} \\
    {} &  Original &  KNN-Feats-3 &  KNN-Role-3 & KNN-Global-3 &  Original &  KNN-Feats-3 &  KNN-Role-3 & KNN-Global-3 \\
    \hline
    Texas     &  2.377261 &    0.761384 &    0.846998 &      0.870676 &  0.107692 &    0.557836 &    0.553922 &      0.574359 \\
    Wisconsin &  1.432984 &    0.580346 &    0.731740 &      0.763613 &  0.196117 &    0.567797 &    0.433272 &      0.399232 \\
    Cornell   &  1.824105 &    0.843352 &    0.981786 &      1.056468 &  0.130872 &    0.552830 &    0.459906 &      0.405063 \\
    Actor     &  2.192559 &    1.256503 &    1.250266 &      1.261272 &  0.219341 &    0.226190 &    0.214680 &      0.215401 \\
    Chameleon &  1.976782 &    1.546773 &    0.912755 &      0.578983 &  0.235007 &    0.227978 &    0.431680 &      0.651369 \\
    Squirrel  &  2.299290 &    1.590335 &    1.017435 &      0.662131 &  0.223943 &    0.213029 &    0.335654 &      0.551077 \\
    \end{tabular}
    \caption{Total variation~\eqref{eq:total_var} and edge homophily~\eqref{eq:edge_homoph} of the node labels $\bby$ measured using four different graphs: (i) the original graph $\ccalG$ from the dataset and three KNN graphs based on (ii) the original node features for $\ccalG_{\text{feat}}$, (iii) role-based structural features for $\ccalG_{\text{role}}$ or (iv) global structural features for $\ccalG_{\text{global}}$. \vspace{-0.5cm}}
    \label{tab:smoothness}
\end{table*}

\begin{itemize}
    \item[(1)] For real-world graphs with heterophilic data, we demonstrate that graphs computed from structural characteristics exhibit greater homophily than the original graph.

    \item[(2)] We present a simple yet adaptable GNN that learns how to combine different representations of the same graph based on role-based or global structural features.
    
    \item[(3)] We empirically demonstrate that considering graphs based on structural features can improve node classification performance on benchmark heterophilic datasets.
    The improvement is even greater when using our proposed GNN to integrate multiple graph representations.

\end{itemize}

\section{Preliminaries and problem statement}
\noindent
\textbf{Fundamentals of GSP.}
Let $\ccalG = (\ccalV, \ccalE)$ be a possibly directed graph, where $\ccalV$ denotes a set of $N$ nodes and $\ccalE \subseteq \ccalV \times \ccalV$ a set of edges, where $(i,j) \in \ccalE$ exists if and only if there is a link from node $j$ to node $i$.
For graphical operations, the connectivity of $\ccalG$ can be conveniently encoded in the adjacency matrix $\bbA \in \reals^{N\times N}$, a potentially weighted matrix with $A_{ij} \neq 0$ if and only if $(j,i) \in \ccalE$.
In addition to the graph $\ccalG$, we assume that we observe signals defined on the set of nodes $\ccalV$.
More specifically, a graph signal is a vector $\bbx \in \reals^N$, where the entry $x_i$ denotes the signal value at node $i$.

\vspace{2mm}
\noindent \textbf{Convolutional GNNs.}
A GNN can be formulated as a parametric non-linear function $f_{\bbTheta}(\bbX | \ccalG)$, which leverages the structure of the underlying graph for downstream predictions.
Among the various GNN architectures, a prominent class relies on an \emph{aggregation function} inspired by graph convolutions.
A popular example within this class is the graph convolutional network (GCN)~\cite{kipf17gnns}, whose output at the layer $\ell$ is given by
\begin{equation}
    \bbX^{(\ell+1)} = \sigma \left( \tbD^{-\frac{1}{2}} \tbA  \tbD^{-\frac{1}{2}} \bbX^{(\ell)} \bbTheta^{(\ell)} \right).
\end{equation}
Here, $\tbA = \bbI + \bbA$ and $\tbD = \diag(\tbA \bbone)$ with $\bbone$ being the vector of all ones and $\diag(\cdot)$ the diagonal operator that converts a vector into a diagonal matrix.
The matrix $\bbTheta^{(\ell)} \in \reals^{F^{(\ell)} \times F^{(\ell+1)}}$ collects the learnable parameters, and $F^{(\ell)}$ and $F^{(\ell+1)}$ are the number of input and output features of the $\ell$-th layer.
The point-wise non-linear activation function $\sigma$ is commonly chosen as the ReLU function, defined as $\sigma(x) = \max(0, x)$.
Despite its simplicity and effectiveness, the GCN architecture is known to struggle under certain scenarios due to its inherent low-pass nature~\cite{zheng2024graphneuralnetworks}.
Differently,~\cite{ruiz2021graph} uses a learnable bank of graph filters to aggregate signals from multiple hops to overcome the homophily assumption. 
In this approach, the output of the $\ell$-th layer is computed as
\begin{equation}\label{eq:fb_gnn}
    \bbX^{(\ell+1)} = \sigma \left( \sum_{r=0}^{R-1} \bbA^r \bbX^{(\ell)} \bbTheta_r^{(\ell)} \right).
\end{equation}
where $\bbTheta_r^{(\ell)} \in \reals^{F^{(\ell)} \times F^{(\ell+1)}}$ collects the learnable parameters and the power $\bbA^r$ determine the radius of the aggregation neighborhood.

\vspace{2mm}
\noindent\textbf{Problem statement.}
We focus on solving the task of semi-supervised node classification; however, our ensuing approach also applies to other graph-based tasks.
Consider a data matrix $\bbX := [\bbx_1,\dots,\bbx_F] \in \reals^{N \times F}$ of $F$ graph signals to be exploited for nodal predictions.
Let $\ccalY := \{y_1,\dots,y_N\}$ represent a set of node labels, of which we only know a subset $\ccalY_{\text{train}} \subset \ccalY$ containing $M$ labels.
The node classification task aims to fit the parameters of a GNN for predicting node classes, that is, to use $f_{\bbTheta}$ to predict unknown labels $\ccalY \backslash \ccalY_{\text{train}}$.
To this end, we solve the following optimization problem
\begin{equation}\label{eq:classical_GNN}
    \min_{\bbTheta} \ccalL \big(f_{\bbTheta}(\bbX | \ccalG), \ccalY_{\text{train}}\big),
\end{equation}
where $f_{\bbTheta}(\bbX | \ccalG)$ represents the output of the GNN, $\bbTheta$ collects the learnable parameters, and $\ccalL$ denotes an appropriate loss function, such as cross-entropy loss.

In many cases, the nodal data such as labels may not be homophilic with respect to the given graph $\ccalG$~\cite{suresh2021breaking}. 
However, the structure in $\ccalG$ may still be semantically relevant~\cite{lee2019graph}. 
Thus, we require an approach to exploit graph structure for downstream node classification.
To this end, we perform neighborhood discovery to obtain a set of alternative graphs based on the original $\ccalG$, and we use all graphs for an adaptive GNN architecture that learns which connections are most relevant.

\begin{figure*}[t!]
    \centering
    \includegraphics[width=.9\textwidth]{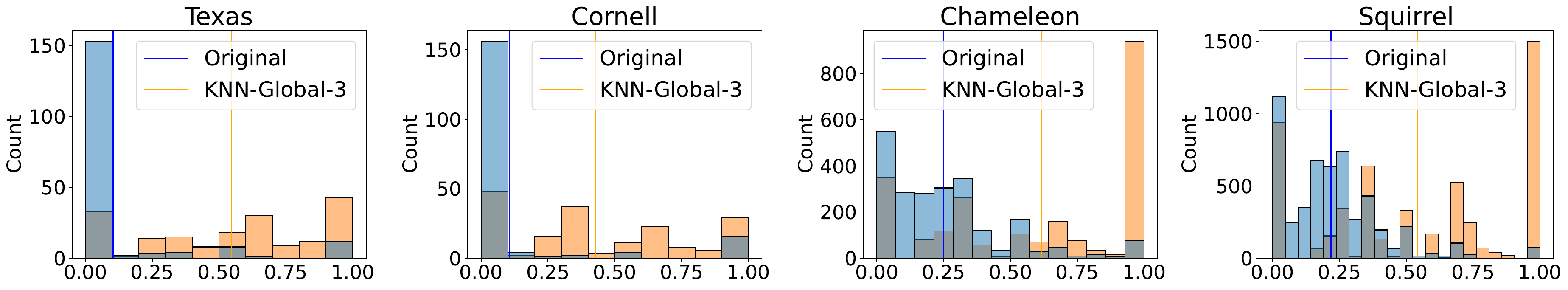}
    \vspace{-0.1cm}
    \caption{Histograms for the node homophilies~\eqref{eq:node_homoph} for 4 graph datasets, as measured using the original graph from the dataset and our KNN graph using global structural features. \vspace{-0.6cm}}
    \label{fig:homophilies}
\end{figure*}

\begin{figure}[b!]
    \centering
    \includegraphics[width=.45\textwidth]{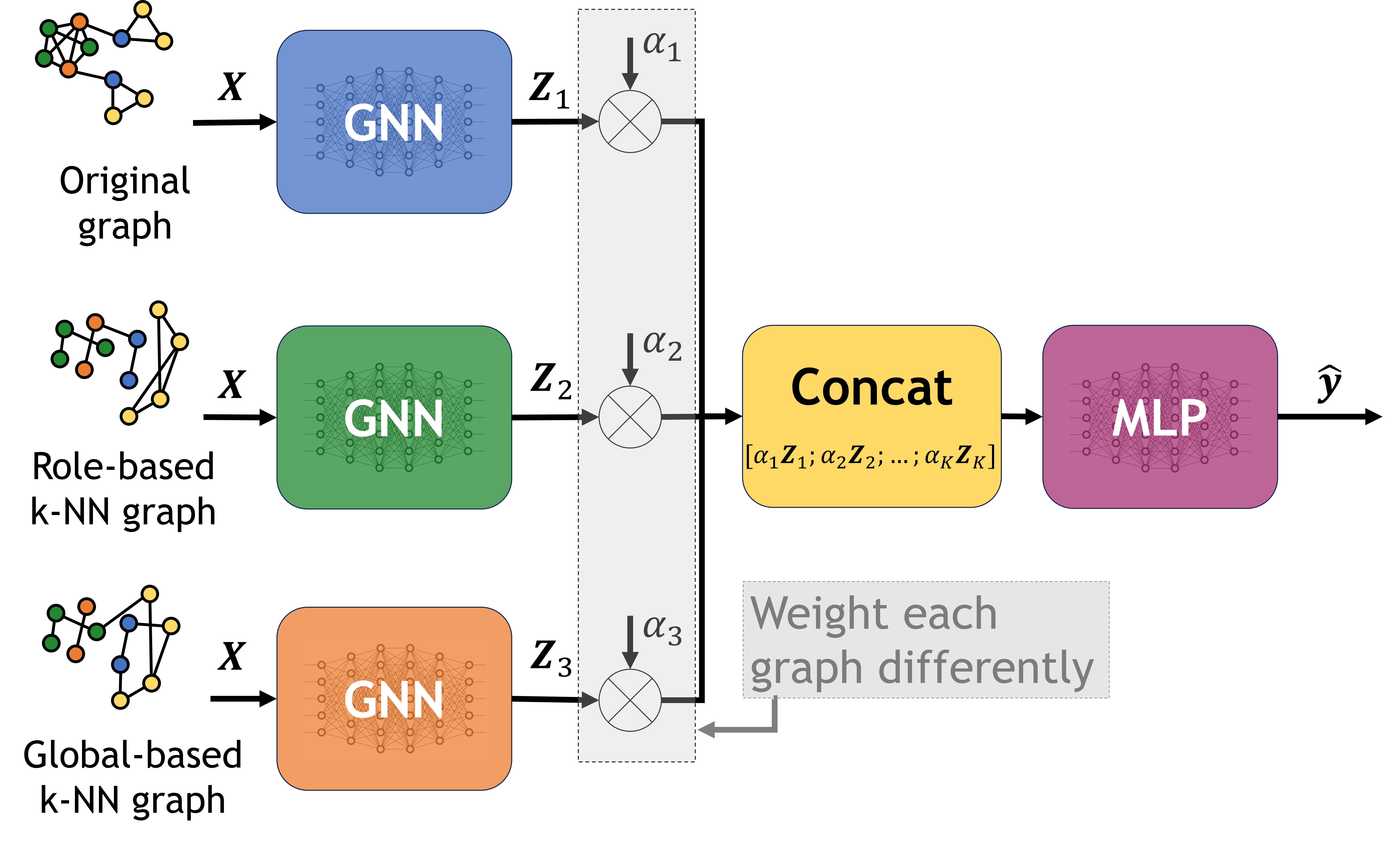}
    \vspace{-0.1cm}
    \caption{Schema of the proposed architecture. 
    }
    \label{fig:gnn}
\end{figure}

\section{Structure-guided GNN}

Our approach follows two steps: (i) Compute alternative graphs based on structural features of each node, then (ii) train an adaptive GNN to combine information from all graphs, i.e., the original graph and the alternative structures.

\vspace{2mm}
\noindent\textbf{Homophily measures.}
The crux of our approach is the computation of the additional graphs.
We intend to obtain new connections that satisfy the homophily assumption.
To this end, we consider the following metrics to measure the homophily of a graph.
A classical approach is to measure the smoothness of the node labels via total variation~\cite{sandryhaila2014discrete}
\begin{equation} \label{eq:total_var}
    TV(\bby) = \| \bby - \bbA\bby \|_1,
\end{equation}
where $\bby \in \reals^N$ collects the values in $\ccalY$.
However, as we are interested in node classification, it is esential to measure the relationship between node labels and graph connections.
Thus, we test if edges indicate shared classes via edge homophily~\cite{zhu2020beyond}
\begin{equation} \label{eq:edge_homoph}
    h_{edge} = \frac{| \{ (i,j) \in \mathcal{E} : y_i = y_j \} | }{|\mathcal{E}|},
\end{equation}
that is, the ratio of edges connecting nodes with the same class label.
Additionally, we may also measure node homophily~\cite{pei2020geom} 
\begin{equation} \label{eq:node_homoph}
    h_{node} (i) = \frac{| \{ j \in \ccalN (i) : y_i = y_j \} | }{|\ccalN (i)|},
\end{equation}
where $\ccalN (i)$ denotes the neighborhood of node $i$.
This metric defines the ratio of the node's neighbors that share its label.
In the sequel, we apply these metrics on real-world data to show that it can exhibit homophily on alternative structures.

\begin{figure*}[t!]
    \centering
    \includegraphics[width=0.9\textwidth]{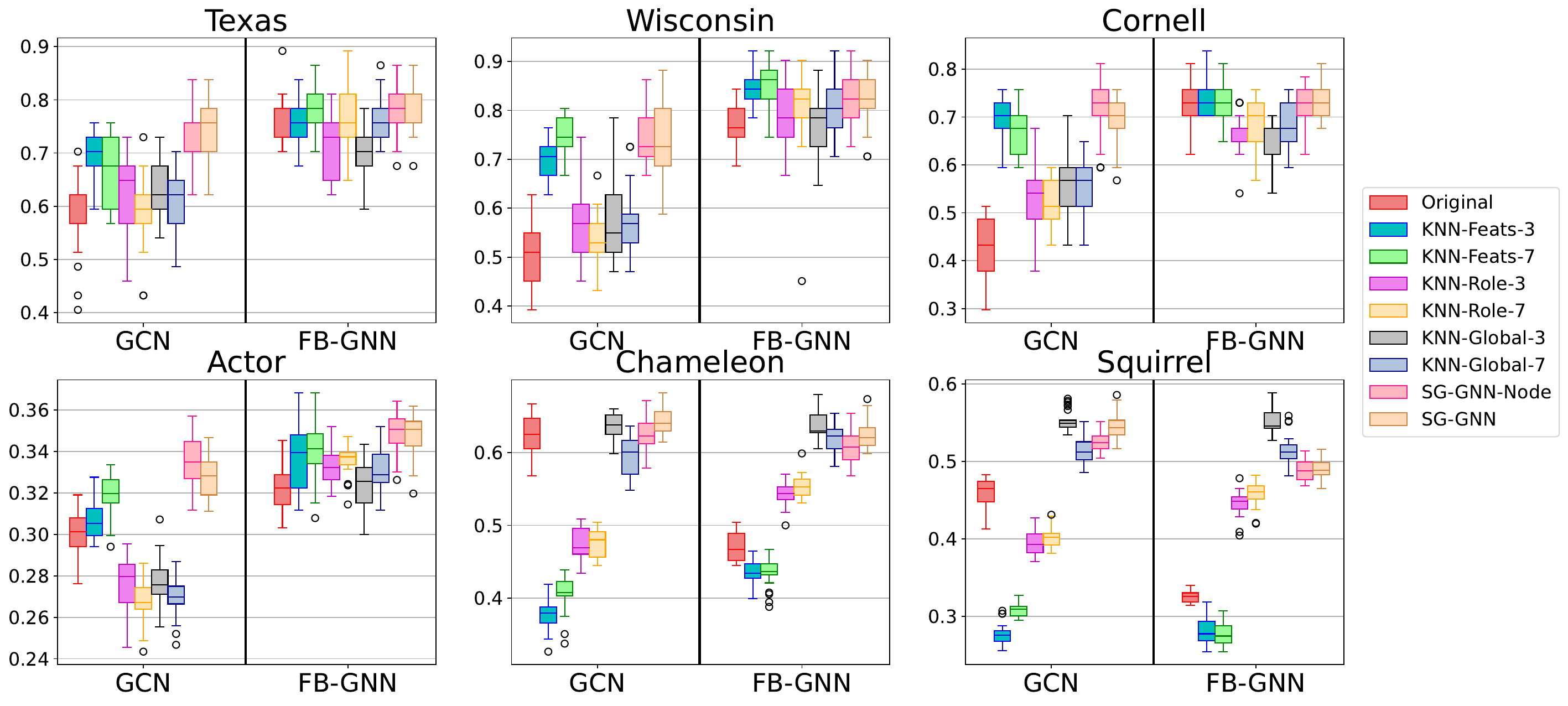}
    \vspace{-0.3cm}
    \caption{Node classification accuracy using the original graph $\ccalG$ and the proposed KNN graphs, $\ccalG_{\text{feat}}$, $\ccalG_{\text{role}}$, and $\ccalG_{\text{global}}$.
    The numbers in the figure legends represent the number of neighbors $k$ used to create each KNN graph. 
    Each boxplot is created for 10 different realizations of train-test splits.\vspace{-0.4cm}}
    \label{fig:exps}
\end{figure*}

\vspace{2mm}
\noindent\textbf{Structure-based neighborhood discovery.}
We operate under the assumption that two nodes are related if they exhibit structural similarity, meaning nodes with similar structural information are likely to share the same labels.
We represent node structural information via two types of features: role-based and global.
The former aims to represent the role a node plays in a graph by considering node-level connectivity patterns, such as the number of triangles in which the node participates or characteristics of its egonet~\cite{guo20role,rossi2015role,gilpin2013guided}. 
The latter characterizes a node's structural characteristics according to its position in the graph by using features such as node eccentricity or centrality measures, often used for ranking nodes~\cite{salavati2019ranking,sheng2020identifying,segarra2015stability}. 

For each node in $\ccalV$, we compute a set of role-based and global structural features based on $\ccalG$, respectively denoted $\bbZ_{\text{role}}$ and $\bbZ_{\text{global}}$\footnote{Our choice of features along with other simulation details can be found at the GitHub repository~\url{https://github.com/vmtenorio/sg-gnn}.}.
Then, we compute two KNN graphs $\ccalG_{\text{role}}$ and $\ccalG_{\text{global}}$ corresponding respectively to $\bbZ_{\text{role}}$ and $\bbZ_{\text{global}}$ by computing the Euclidean distances between the structural features for every pair of nodes.
More specifically, each new graph $\ccalG_{\text{role}}$ or $\ccalG_{\text{global}}$ comprises the same set of nodes $\ccalV$ as the original graph $\ccalG$, but each node will be connected to its $k$-nearest neighbors with respect to distances between role-based or global structural features.
If these characteristics are correlated with node labels, then they will be homophilic on $\ccalG_{\text{role}}$ or $\ccalG_{\text{global}}$.

We next demonstrate that this approach is feasible for real-world data.
In particular, for multiple graph datasets, we use the metrics ~\eqref{eq:total_var},~\eqref{eq:edge_homoph}, and~\eqref{eq:node_homoph} to measure homophily with respect to the original graph $\ccalG$ versus three alternative KNN graphs using $k=3$ neighbors: the role-based $\ccalG_{\text{role}}$, the global $\ccalG_{\text{global}}$, and a third alternative $\ccalG_{\text{feat}}$ based on distances between node features $\bbX$.
In Table~\ref{tab:smoothness}, we measure the total variation~\eqref{eq:total_var} and edge homophily~\eqref{eq:edge_homoph} for all four graphs on six datasets.
For all datasets, total variation is higher and edge homophily lower on $\ccalG$ versus all alternatives.
Moreover, while $TV$ is lower for $\ccalG_{\text{feat}}$ for the Texas, Wisconsin, and Cornell datasets, the role-based and global graphs $\ccalG_{\text{role}}$ and $\ccalG_{\text{global}}$ show smoother behavior via $TV$ for the Actor, Chameleon, and Squirrel datasets.
Similarly, while $\ccalG_{\text{feat}}$ has higher $h_{edge}$ for the Wisconsin, Cornell, and Actor datasets, $\ccalG_{\text{role}}$ or $\ccalG_{\text{global}}$ attain higher $h_{edge}$ for the Texas, Chameleon, and Squirrel datasets.
Thus, our KNN graphs based on structural features may be more suitable for GNN use than $\ccalG$.
Moreover, in Fig.~\ref{fig:homophilies} we measure node homophily $h_{node}$ for the same  six datasets.
In particular, we plot the histograms of $h_{node}$ computed for each node in $\ccalV$ with respect to $\ccalG$ and $\ccalG_{\text{global}}$.
For the original graph $\ccalG$, $h_{node}$ is concentrated towards lower values, while the global $\ccalG_{\text{global}}$ exhibits greater overall homophily.
As we are interested in node classification, an increase in node homophily motivates the use of these alternative structures for GNN predictions.

\vspace{2mm}
\noindent\textbf{Adaptive GNN.}
Finally, we introduce an adaptive architecture which can exploit more than one graph, namely $I$ graphs, including the original $\ccalG$, the role-based $\ccalG_{\text{role}}$, and the global $\ccalG_{\text{global}}$.
Our approach is shown in Fig.~\ref{fig:gnn}, where each graph is assigned a separate GNN from which we obtain a set of node embeddings $\{\bbZ_i\}_{i=1}^I$. 
We learn the corresponding weights $\{\alpha_i\}_{i=1}^I$ for each set of embeddings so the model can flexibly combine all input graphs.
Finally, the weighted embeddings are concatenated obtaining $\bbZ = [\alpha_1 \bbZ_1,\ldots, \alpha_I \bbZ_I]$, and passed through a multi-layer perceptron (MLP), which outputs the label prediction.
To ensure that the energy is preserved, we enforce $\sum_{i=1}^I \alpha_i = 1$ via a softmax activation function.
Our approach allows us to still exploit the original graph $\ccalG$, and we also learn which alternative structure is most informative.
By learning the weights $\{\alpha_i\}$, we can interpret which type of connections are most relevant to the node classification task.

Bearing in mind the intuition from Fig.~\ref{fig:homophilies}, we note that the nodes behave very differently within the same datasets, with nodes whose entire neighborhood share the same label ($h_{node} (n) = 1$), and others with no neighbor having the same label ($h_{node} (n) = 0$).
To leverage this diversity, we propose to learn node-specific aggregation coefficients $\alpha_i^{n}$, and create the vectors $\{\bbalpha_i\}_{i=1}^I$ with $\bbalpha_i \in \reals^N$. The concatenation in this case is computed as $\bbZ = [\text{diag}(\bbalpha_1) \bbZ_1,\ldots, \text{diag}(\bbalpha_I) \bbZ_I]$ so that the $n$-th row of $\bbZ_i$ is multiplied by $[\alpha_i]_n$. This allows each node to choose the graph from which to perform the aggregation, hopefully leading to an improved performance.
Similarly to the previous case, we enforce that $\sum_{i=1}^I [\alpha_i]_n = 1, \; \forall n \in \ccalV$.

\section{Experiments}

We next numerically evaluate our proposed approach of using the proposed KNN graphs $\ccalG_{\text{role}}$ and $\ccalG_{\text{global}}$ for node classification on heterophilic datasets.
We also demonstrate the viability of our adaptive approach summarized in Fig.~\ref{fig:gnn}.

\vspace{2mm}
\noindent
\textbf{Datasets}.
We evaluate our method using six widely-used heterophilic datasets.
Texas, Cornell, and Wisconsin are subgraphs of the WebKB dataset~\cite{craven1998learning}, where nodes represent webpages from university computer science departments and edges denote hyperlinks.
In the Actor dataset~\cite{pei2020geom} nodes represent actors, and edges indicate co-occurrence on the same Wikipedia page.
In Chameleon and Squirrel~\cite{rozemberczki2021multiscale} nodes are Wikipedia articles while edges represent mutual links.

\vspace{2mm}
\noindent
\textbf{GNN Architectures}.
These experiments aim to (i) validate the hypothesis that using graphs on which labels are smoother improves the performance of low-pass GNN architectures and (ii) assess the impact of alternative graphs and our adaptive architecture using a GNN suitable for heterophilic data.
Consequently, we use the GCN~\cite{kipf17gnns} as a homophilic baseline and the filter-bank GCN (FB-GCN) in~\eqref{eq:fb_gnn} from~\cite{ruiz2021graph} as a non-homophilic baseline.
We also evaluate our adaptive architecture, named SG-GNN, with and without node-specific weights (denoted as ``SG-GNN-Node'' and ``SG-GNN'').

\vspace{2mm}
\noindent
\textbf{Input Graphs}.
We use the original graph and KNN graphs with either $k=3$ or $k=7$ neighbors as indicated in the legend of Fig.~\ref{fig:exps}.
The KNN graphs are: (i) $\ccalG_{\text{feat}}$ based on the node features $\bbX$ (denoted ``Feats''); (ii) $\ccalG_{\text{role}}$ based on role structural features (denoted ``Role''), and (iii) $\ccalG_{\text{global}}$ based on global structural features (denoted ``Global'').

\vspace{2mm}
\noindent
\textbf{Results}.
Fig.~\ref{fig:exps} depicts the test accuracy when using different graphs as input for GCN and FB-GCN.
First, we observe that using the original graph as input is outperformed by at least one of the KNN graphs in all datasets.
Nevertheless, the best-performing graph is not consistent. 
For Texas, Wisconsin, Cornell and Actor datasets, the graph $\ccalG_{\text{feat}}$ (green and blue boxes) shows the highest performance, while in Chameleon and Squirrel, $\ccalG_{\text{feat}}$ offers the worst performance, being the graph $\ccalG_{\text{global}}$ (black and dark blue boxes) the preferred alternative.
Although identifying the preferred graph beforehand is non-trivial, our SG-GNN consistently outperforms the best-performing individual input graph for both the GCN and FB-GCN.
This highlights the capacity of SG-GNN to adaptively learn which graph (or combination thereof) is the most suitable for each dataset.

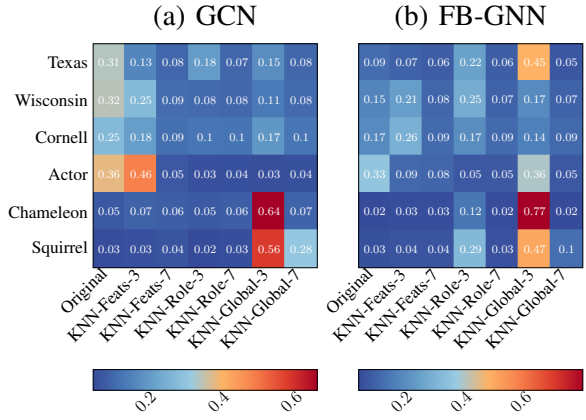
\begin{figure}[t]
    \centering
    \begin{tikzpicture}[baseline,scale=.55]

\pgfplotsset{colormap={CM}{
    rgb255=(54, 75, 154)
    rgb255=(74, 123, 183)
    rgb255=(110, 166, 205)
    rgb255=(152, 202, 225)
    rgb255=(253, 179, 102)
    rgb255=(246, 126, 75)
    rgb255=(221, 61, 45)
    rgb255=(165, 0, 38)
}}

\begin{groupplot}[
    name=learned_coefs,
    table/col sep=space,
    width=7cm,
    height=7cm,
    group style={group size=2 by 1,
        horizontal sep=1cm,
        vertical sep=1.8cm,},
    enlargelimits=false,
    colorbar horizontal,
    xtick={0,1,2,3,4,5,6},
    xticklabels={Original,KNN-Feats-3,KNN-Feats-7,KNN-Role-3,KNN-Role-7,KNN-Global-3,KNN-Global-7},
    x tick label style={rotate=-45,anchor=east},
    ytick={0,1,2,3,4,5},
    yticklabels={Texas,Wisconsin,Cornell,Actor,Chameleon,Squirrel},
    axis on top,
    label style={font=\LARGE},
    tick style={draw=none},
    tick label style={font=\large},
    title style={font=\huge},
    x tick label style={rotate=90}
    ]

    \pgfplotstableread{figures/data/coefs-GCN.csv}\matrixA
    \pgfplotstableread{figures/data/coefs-GCNH.csv}\matrixB

\nextgroupplot[title={(a) GCN},xmin=-0.5,xmax=6.5,ymin=-0.5,ymax=5.5]
    \addplot [
        matrix plot,
        colorbar,
        nodes near coords,
        every node near coord/.append style={
            font=\footnotesize,
            /pgf/number format/fixed,        
            /pgf/number format/precision=2,  
            text=white,
            anchor=center
        },
        point meta=explicit,
        mesh/cols=7,  
    ] table [meta=value] {\matrixA};

\nextgroupplot[title={(b) FB-GNN},xmin=-0.5,xmax=6.5,ymin=-0.5,ymax=5.5,yticklabels={}]
    \addplot [
        matrix plot,
        colorbar,
        nodes near coords,
        every node near coord/.append style={
            font=\footnotesize,
            /pgf/number format/fixed,        
            /pgf/number format/precision=2,  
            text=white,
            anchor=center,
        },
        point meta=explicit,
        mesh/cols=7,  
    ] table [meta=value] {\matrixB};

\end{groupplot}
\end{tikzpicture}
    \caption{Heat maps of coefficients $\{\alpha_i\}$ learned from the proposed SG-GNN architecture.
    }
    \label{fig:learned_coefs}
\end{figure}

We confirm this in the following ablation study where we analyze the coefficients $\{\alpha_i\}$ learned by the SG-GNN architecture, which are represented in Fig.~\ref{fig:learned_coefs}.
Indeed, the graph associated with the highest coefficient and the best performing input graph coincide for every dataset.
For example, the KNN graph with $k=3$ based on global structural features $\ccalG_{\text{global}}$ achieves the highest coefficients for the Chameleon and Squirrel datasets, reflecting its superior accuracy in Fig.~\ref{fig:exps}.
This confirms that our adaptive architecture is able to select the most suitable graph for the underlying task.

\section{Conclusions}

In this work, we presented a neighborhood discovery approach to node classification.
Instead of the original graph, we proposed computing KNN graphs based on either role-based or global structural features.
We demonstrated that these options are potentially better suited to the low-pass filtering operations performed by many GNNs, as we observe an increase in homophilic behavior.
We then proposed an adaptive GNN approach that learns how to combine the original graph with the structure-based KNN graphs to node classification.
Our empirical results confirm that the proposed structure-based KNN graphs are suitable for heterophilic datasets and that our adaptive architecture effectively identifies the most appropriate graph for a given classification task.
In future work, we may consider learning structural features as opposed to using a pre-defined set of role-based or global features.

\bibliographystyle{IEEEbib}
\bibliography{myIEEEabrv,citations}

\begin{thebibliography}{10}

\bibitem{wu2021comprehensive}
Z.~Wu, S.~Pan, F.~Chen, G.~Long, C.~Zhang, and P.~S. Yu,
\newblock ``A comprehensive survey on graph neural networks,''
\newblock {\em IEEE Trans. Neural Netw. and Learning Syst.}, vol. 32, no. 1,
  pp. 4--24, 2021.

\bibitem{chien2024opportunities}
E.~Chien, M.~Li, A.~Aportela, et~al.,
\newblock ``Opportunities and challenges of graph neural networks in electrical
  engineering,''
\newblock {\em Nature Reviews Electrical Engineering}, vol. 1, no. 8, pp.
  529--546, 2024.

\bibitem{zheng2024graphneuralnetworks}
X.~Zheng, Y.~Wang, Y.~Liu, M.~Li, M.~Zhang, D.~Jin, P.~S. Yu, and S.~Pan,
\newblock ``Graph neural networks for graphs with heterophily: A survey,''
\newblock {\em arXiv:2202.07082}, 2024.

\bibitem{chien2022node}
E.~Chien, W.~Chang, C.~Hsieh, H.~Yu, J.~Zhang, O.~Milenkovic, and I.~S.
  Dhillon,
\newblock ``Node feature extraction by self-supervised multi-scale neighborhood
  prediction,''
\newblock in {\em Intl. Conf. Learning Representations (ICLR)}, 2022.

\bibitem{suresh2021breaking}
S.~Suresh, V.~Budde, J.~Neville, P.~Li, and J.~Ma,
\newblock ``Breaking the limit of graph neural networks by improving the
  assortativity of graphs with local mixing patterns,''
\newblock in {\em Intl. Conf. Knowledge Discovery \& Data Mining (SIGKDD)},
  2021, pp. 1541--1551.

\bibitem{jin2020gralsp}
Y.~Jin, G.~Song, and C.~Shi,
\newblock ``{{GraLSP}}: Graph neural networks with local structural patterns,''
\newblock {\em AAAI Conf. Artif. Intell.}, vol. 34, no. 04, pp. 4361--4368,
  2020.

\bibitem{long2020graph}
Q.~Long, Y.~Jin, G.~Song, Y.~Li, and W.~Lin,
\newblock ``Graph structural-topic neural network,''
\newblock in {\em Intl. Conf. Knowledge Discovery \& Data Mining (SIGKDD)},
  2020, pp. 1065--1073.

\bibitem{lee2019graph}
J.~B. Lee, R.~A. Rossi, X.~Kong, S.~Kim, E.~Koh, and A.~Rao,
\newblock ``Graph convolutional networks with motif-based attention,''
\newblock in {\em Intl. Conf. Knowledge Discovery \& Data Mining (SIGKDD)},
  2019, pp. 499--508.

\bibitem{xu2018representation}
K.~Xu, C.~Li, Y.~Tian, T.~Sonobe, K.~Kawarabayashi, and S.~Jegelka,
\newblock ``Representation learning on graphs with jumping knowledge
  networks,''
\newblock in {\em Intl. Conf. Machine Learning (ICML)}. 2018, vol.~80 of {\em
  Proc. Mach. Learn. Res.}, pp. 5453--5462, PMLR.

\bibitem{ruiz2021graph}
L.~Ruiz, F.~Gama, and A.~Ribeiro,
\newblock ``Graph neural networks: Architectures, stability, and
  transferability,''
\newblock {\em Proc. {IEEE}}, vol. 109, no. 5, pp. 660--682, 2021.

\bibitem{rey2024redesigning}
S.~Rey, M.~Navarro, V.~M. Tenorio, S.~Segarra, and A.~G. Marques,
\newblock ``Redesigning graph filter-based {GNNs} to relax the homophily
  assumption,''
\newblock {\em arXiv preprint arXiv:2409.08676}, 2024.

\bibitem{liu2022towards}
Y.~Liu, Y.~Zheng, D.~Zhang, H.~Chen, H.~Peng, and S.~Pan,
\newblock ``Towards unsupervised deep graph structure learning,''
\newblock in {\em Proc. ACM Web Conf.}, 2022, pp. 1392–--1403.

\bibitem{yang2021consisrec}
L.~Yang, Z.~Liu, Y.~Dou, J.~Ma, and P.~S. Yu,
\newblock ``{ConsisRec}: Enhancing {GNN} for social recommendation via
  consistent neighbor aggregation,''
\newblock in {\em Proc. Intl. ACM SIGIR Conf. Res. Devel. Inf. Retr.}, 2021,
  pp. 2141–--2145.

\bibitem{balcilar2021analyzingexpressivepower}
M.~Balcilar, G.~Renton, P.~H{\'e}roux, B.~Ga{\"u}z{\`e}re, S.~Adam, and
  P.~Honeine,
\newblock ``Analyzing the expressive power of graph neural networks in a
  spectral perspective,''
\newblock in {\em Intl. Conf. Learning Representations (ICLR)}, 2021.

\bibitem{tenorio2024recovering}
V.~M. Tenorio, M.~Navarro, S.~Segarra, and A.~G. Marques,
\newblock ``Recovering missing node features with local structure-based
  embeddings,''
\newblock in {\em IEEE Intl. Conf. Acoust., Speech and Signal Process.
  (ICASSP)}. IEEE, 2024, pp. 9931--9935.

\bibitem{craven1998learning}
M.~Craven, D.~DiPasquo, D.~Freitag, A.~McCallum, T.~Mitchell, K.~Nigam, and
  S.~Slattery,
\newblock ``Learning to extract symbolic knowledge from the {World Wide Web},''
\newblock in {\em AAAI Conf. Artif. Intell.}, 1998, p. 509–516.

\bibitem{salavati2019ranking}
C.~Salavati, A.~Abdollahpouri, and Z.~Manbari,
\newblock ``Ranking nodes in complex networks based on local structure and
  improving closeness centrality,''
\newblock {\em Neurocomputing}, vol. 336, pp. 36--45, 2019.

\bibitem{sheng2020identifying}
J.~Sheng, J.~Dai, B.~Wang, G.~Duan, J.~Long, J.~Zhang, K.~Guan, S.~Hu, L.~Chen,
  and W.~Guan,
\newblock ``Identifying influential nodes in complex networks based on global
  and local structure,''
\newblock {\em Physica A: Stat. Mechanics and its Applications}, vol. 541, pp.
  123262, 2020.

\bibitem{segarra2015stability}
S.~Segarra and A.~Ribeiro,
\newblock ``Stability and continuity of centrality measures in weighted
  graphs,''
\newblock {\em IEEE Trans. Signal Process.}, vol. 64, no. 3, pp. 543--555,
  2015.

\bibitem{peng2021reinforced}
H.~Peng, R.~Zhang, Y.~Dou, R.~Yang, J.~Zhang, and P.~S. Yu,
\newblock ``Reinforced neighborhood selection guided multi-relational graph
  neural networks,''
\newblock {\em ACM Trans. Inf. Syst.}, vol. 40, no. 4, 2021.

\bibitem{kipf17gnns}
T.~N. Kipf and M.~Welling,
\newblock ``Semi-supervised classification with graph convolutional networks,''
\newblock in {\em Intl. Conf. Learning Representations (ICLR)}, 2017.

\bibitem{sandryhaila2014discrete}
A.~Sandryhaila and J.~M.~F. Moura,
\newblock ``Discrete signal processing on graphs: Frequency analysis,''
\newblock {\em IEEE Trans. Signal Process.}, vol. 62, no. 12, pp. 3042--3054,
  2014.

\bibitem{zhu2020beyond}
J.~Zhu, Y.~Yan, L.~Zhao, M.~Heimann, L.~Akoglu, and D.~Koutra,
\newblock ``Beyond homophily in graph neural networks: Current limitations and
  effective designs,''
\newblock in {\em Conf. Neural Inform. Process. Syst.}, 2020.

\bibitem{pei2020geom}
H.~Pei, B.~Wei, K.~C.-C. Chang, Y.~Lei, and B.~Yang,
\newblock ``Geom-{GCN}: Geometric graph convolutional networks,''
\newblock {\em arXiv preprint arXiv:2002.05287}, 2020.

\bibitem{guo20role}
X.~Guo, W.~Zhang, W.~Wang, Y.~Yu, Y.~Wang, and P.~Jiao,
\newblock ``Role-oriented graph auto-encoder guided by structural
  information,''
\newblock in {\em Database Syst. Advanced Appl.}, 2020, pp. 466--481.

\bibitem{rossi2015role}
R.~A. Rossi and N.~K. Ahmed,
\newblock ``Role discovery in networks,''
\newblock {\em IEEE Trans. Knowledge and Data Engineering}, vol. 27, no. 4, pp.
  1112--1131, 2015.

\bibitem{gilpin2013guided}
S.~Gilpin, T.~Eliassi-Rad, and I.~Davidson,
\newblock ``Guided learning for role discovery ({GLRD}): framework, algorithms,
  and applications,''
\newblock in {\em Intl. Conf. Knowledge Discovery \& Data Mining (SIGKDD)},
  2013, KDD '13, p. 113–121.

\bibitem{rozemberczki2021multiscale}
B.~Rozemberczki, C.~Allen, and R.~Sarkar,
\newblock ``Multi-scale attributed node embedding,''
\newblock {\em J. Complex Netw.}, vol. 9, no. 2, pp. cnab014, 2021.

\end{thebibliography}

\end{document}